\documentclass[lettersize,journal]{IEEEtran}
\usepackage{amsmath,amsfonts}
\usepackage{algorithmic}
\usepackage{algorithm}
\usepackage{array}
\usepackage[caption=false,font=normalsize,labelfont=sf,textfont=sf]{subfig}
\usepackage{textcomp}
\usepackage{stfloats}
\usepackage{url}
\usepackage{verbatim}
\usepackage{graphicx}
\usepackage{cite}
\begin{document}

\title{A Multiplication-Free Spike-Time Learning Algorithm and its Efficient FPGA Implementation for On-Chip SNN Training}

\author{Maryam Mirsadeghi, Mojtaba Mirbagheri, Saeed Reza Kheradpisheh}



\maketitle

\begin{abstract}
Spiking Neural Networks (SNNs) offer a biologically inspired foundation for low-power, event-driven intelligence, yet their direct on-chip supervised training remains a key hardware challenge. This paper presents a multiplication-free, spike-time-based learning algorithm specifically designed for efficient FPGA realization. The proposed approach eliminates floating-point arithmetic and explicit gradient storage, enabling a fully event-driven, digital training pipeline. Implemented on a Xilinx Artix-7 FPGA, the architecture achieves high operating speed and minimal resource usage while maintaining competitive accuracy. These results demonstrate that the learning algorithm effectively maps onto reconfigurable hardware, achieving both computational and energy efficiency. Software simulations further validate scalability, with 96.5\% and 84.8\% accuracy on MNIST and Fashion-MNIST. With its spike-driven and multiplier-free operation, the proposed framework delivers a practical and scalable hardware solution for real-time, on-chip SNN learning in edge environments.

\end{abstract}


\section{Introduction}
\IEEEPARstart{S}piking Neural Networks (SNNs) have emerged as a biologically inspired and hardware-efficient alternative to artificial neural networks (ANNs), offering event-driven computation and sparse information processing that closely resemble biological neurons~\cite{p12,p13}. Their ability to represent and process information through spike timing enables low-power, low-latency, and temporally sparse operation—key advantages for edge and neuromorphic systems. Among various neural coding strategies, single-spike temporal coding has received particular attention for its compactness, simplicity, and suitability for hardware implementation~\cite{p14,p15,p16,p17,p18}. In this scheme, each neuron emits at most one spike whose latency carries information, drastically reducing spike activity while maintaining competitive accuracy on benchmark tasks~\cite{temp1,temp2,p18}.

Despite these promising properties, direct on-chip supervised training of SNNs remains a major challenge. Traditional computing architectures such as GPUs or CPUs, based on the Von-Neumann model, are inefficient for temporal, event-driven operations due to their sequential execution and high arithmetic overhead~\cite{trunorth}. While neuromorphic processors such as Intel's Loihi~\cite{loihi}, IBM's TrueNorth~\cite{trunorth}, SpiNNaker~\cite{spinnaker}, and SynSense's DYNAP~\cite{dynap} achieve impressive energy efficiency for inference, they offer limited or no support for supervised, online learning—particularly in deep SNNs. This limitation has motivated research toward reconfigurable and parallel hardware platforms such as Field-Programmable Gate Arrays (FPGAs), which combine high-speed digital processing with low power consumption and offer flexibility for realizing both inference and learning on device~\cite{hasist,online1}.

Early on-chip learning strategies in SNNs predominantly relied on unsupervised Hebbian rules such as spike-timing-dependent plasticity (STDP)~\cite{p19,p20,p21,p22,p23,p24,p25,p26}. Although STDP-based learning is simple to implement and highly local, its lack of supervision and limited scalability hinder its performance on complex tasks~\cite{p2}. Supervised learning, in contrast, enables SNNs to achieve competitive accuracy with ANNs but often comes with significant computational cost. Rate-based supervised models, typically trained using surrogate gradient (SG) learning~\cite{p3,p4,p5}, approximate the spiking function's derivative to enable backpropagation. While effective for training deep networks, rate coding produces dense spike trains and relies on floating-point multiplications, limiting suitability for energy-constrained hardware. 

Temporal coding offers a more hardware-efficient alternative by encoding information in the precise timing of spikes rather than firing rates. In single-spike temporal models such as SpikeProp~\cite{p6} and more recent frameworks like S4NN, BS4NN, and related approaches~\cite{p9,p15,p16,p17,p18,Mostafa_1}, neurons emit at most one spike, with learning driven by spike-time errors between target and actual firing times. These algorithms achieve competitive accuracy with minimal spiking activity but still depend on floating-point arithmetic and gradient storage, making them challenging to implement directly on digital neuromorphic hardware. Multi-spike temporal models and Backpropagation Through Time (BPTT)~\cite{p30,p31,p32,p33,p35,p36} have been explored to handle temporal dependencies but require sequential unrolling, large memory, and real-valued gradient computations, which are incompatible with efficient on-chip learning.

From a hardware perspective, the energy efficiency of SNNs has spurred the development of dedicated neuromorphic chips such as Loihi~\cite{loihi}, TrueNorth~\cite{trunorth}, and SpiNNaker~\cite{spinnaker}. These platforms excel at inference but have limited capability for supervised on-device learning. Most existing hardware implementations rely on offline training, where SNNs are trained on GPUs using surrogate gradients or BPTT, and the trained weights are later transferred for inference~\cite{off1,off2}. While this approach simplifies design, it prevents real-time adaptation and hinders deployment in edge applications requiring online learning. 

Recent works have attempted to bring supervised training into hardware by leveraging FPGA and ASIC platforms for their reconfigurability and parallelism~\cite{hasist,online1}. For instance, HaSiST~\cite{hasist} introduced an FPGA-based supervised learning engine that performs event-driven processing; however, it uses rate-based coding and floating-point multipliers, leading to high hardware overhead and limited scalability. Similarly, other FPGA implementations rely on hybrid CPU–FPGA frameworks or partial floating-point computation~\cite{online1}, which increases complexity and power consumption. These limitations underscore the urgent need for fully event-driven, multiplier-free learning frameworks that can achieve real-time on-chip learning while maintaining high hardware efficiency.

To address these challenges, this paper proposes a \textit{multiplication-free, spike-time-based supervised learning algorithm} that unifies algorithmic simplicity with hardware efficiency. Unlike conventional gradient-based or STDP-inspired rules, the proposed approach encodes gradient information entirely in spike latencies, eliminating floating-point multiplications and explicit gradient storage. Both the forward and backward passes operate solely via discrete spike events, transforming the training process into an event-driven flow suitable for FPGA implementation. The design leverages fixed-point arithmetic, parallel adder-based computation, and sparse, distributed spike communication to enable low-power, high-speed on-chip learning. 

This framework tightly couples the algorithmic behavior of SNNs with the hardware architecture, achieving high throughput and minimal resource utilization while maintaining learning capability comparable to SG-based methods and existing FPGA implementations such as HaSiST. To the best of our knowledge, this is among the first works to realize fully spike-driven supervised learning for deep SNNs on FPGA without floating-point arithmetic. The presented algorithm and architecture together bridge the gap between biologically inspired computation and practical neuromorphic systems, paving the way for scalable, real-time on-chip learning in energy-constrained edge environments.

\section{The Proposed Method}
In this section, we present our proposed spiking neural network (SNN) model and its supervised training algorithm tailored for on-chip implementation. The network employs single-spike temporal coding for forward inference and introduces a novel spike-time-driven backpropagation mechanism, eliminating floating-point multiplications and enabling efficient online training using only spike timings. (See Figure\ref{algorithm} for an overview of the proposed SNN with forward and backward spike-time propagation.)

\begin{figure}[!t]
\centering
\includegraphics[width=\linewidth]{Algorithm.jpg}
\caption{Proposed SNN with forward/backward spike-time propagation. Input image is encoded into spikes, processed by IF neurons, and classified by the earliest output spike. Backward spikes propagate temporal gradients during training.}
\label{algorithm}
\end{figure}

\subsection{Network Architecture}
The proposed SNN is a multi-layer feedforward architecture composed of non-leaky integrate-and-fire (IF) neurons. Each neuron emits at most one spike per input sample, with information encoded in spike latency. Input images are converted into spike trains via intensity-to-latency coding. Intermediate layers process these spikes to support efficient on-chip supervised learning without multiplication operations. The output layer consists of neurons corresponding to class labels, where the earliest-firing neuron determines the classification decision.
    
\subsection{Forward Pass}
The forward pass converts input images into spike trains using single-spike temporal coding. For an input image with pixel intensities in the range $[0, I_{\text{max}}]$, the firing time of the $i$-th input neuron, $t_i$, is computed as:
\begin{equation}
t_i = \left\lfloor \frac{I_{\text{max}} - I_i}{I_{\text{max}}} \cdot t_{\text{max}} \right\rfloor
\label{eq1}
\end{equation}
where $t_{\text{max}}$ is the maximum simulation time. Higher pixel intensities result in earlier spikes, ensuring efficient encoding. The spike train for the $i$-th input neuron is:
\begin{equation}
S^0_i(t) =
\begin{cases}
1 & \text{if } t = t_i \\
0 & \text{otherwise}
\end{cases}
\end{equation}
This defines a discrete-time spike at time $t_i$.

\begin{figure}[!t]
\centering
\includegraphics[width=\linewidth]{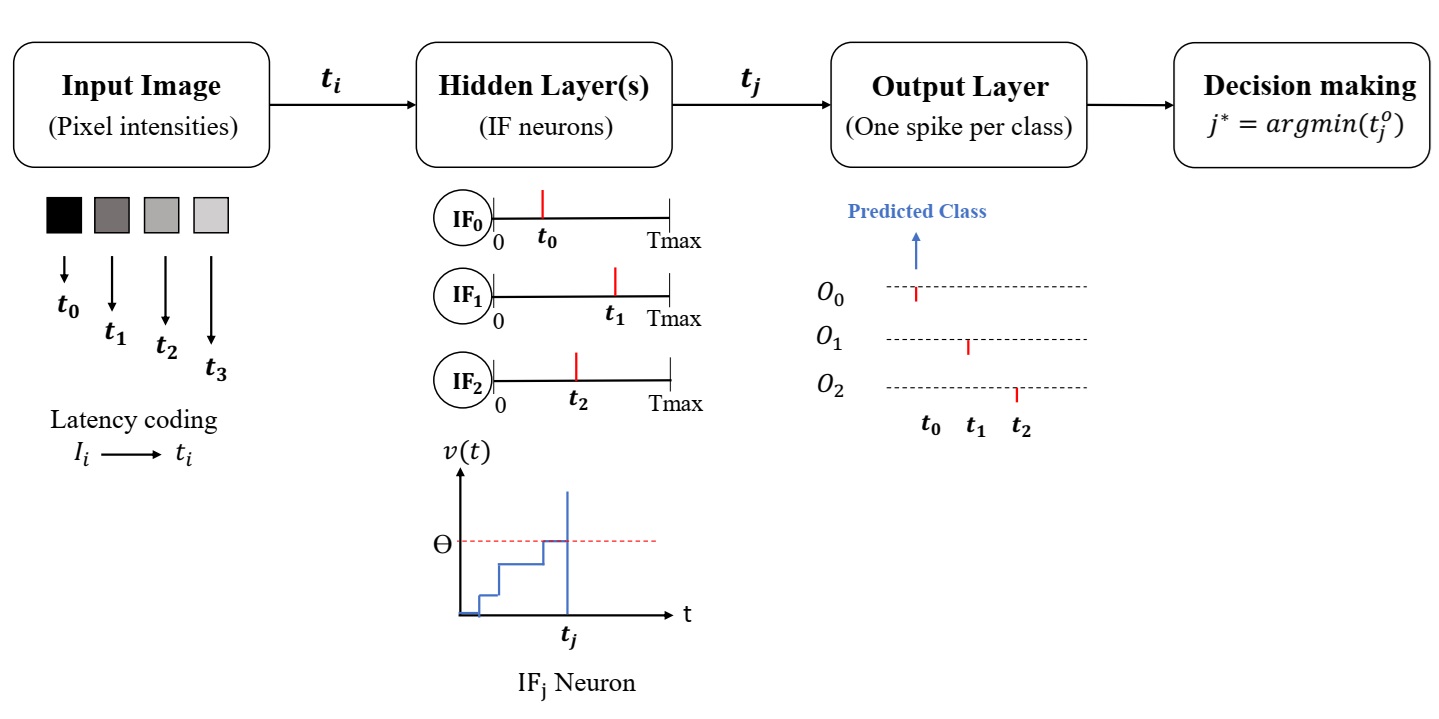}
\caption{Forward pass of the proposed SNN. Input is latency-coded into spike times $t_i$ and propagated through IF neurons. The earliest output spike determines the predicted class.}
\label{forward_alg}
\end{figure}

In hidden and output layers, non-leaky IF neurons update their membrane potential $V^l_j(t)$ based on incoming spikes:
\begin{equation}
V^l_j(t) = V^l_j(t-1) + \sum_i W^l_{ji} S^{l-1}_i(t)
\end{equation}
where $W^l_{ji}$ is the synaptic weight from the $i$-th presynaptic neuron in layer $l-1$ to neuron $j$ in layer $l$, and $S^{l-1}_i(t)$ is the input spike train. A neuron fires once when its membrane potential exceeds the threshold $\theta^l_j$:
\begin{equation}
S^l_j(t) =
\begin{cases}
1 & \text{if } V^l_j(t) \geq \theta^l_j \;\text{and}\; S^l_j(<t) \neq 1 \\
0 & \text{otherwise}
\end{cases}
\end{equation}
The condition $S^l_j(<t) \neq 1$ ensures a neuron fires only once.
Figure~\ref{forward_alg} illustrates this forward-pass mechanism, showing how an input is temporally encoded and how a hidden IF neuron accumulates input spikes to generate a single output spike when the threshold is crossed.

For each input, membrane potentials are reset to zero, and the simulation runs for at most $t_{\text{max}}$ time steps. In the test phase, the first output neuron to fire determines the class. If no output neuron fires by $t_{\text{max}}$, the neuron with the highest membrane potential is selected. During training, silent neurons emit a placeholder spike at $t_{\text{max}}$ to enable error computation.

\subsection{Backward pass}

For a categorization task with $C$ categories, we define the temporal error as a function of the actual and target firing times,
\begin{equation}
e = [e_1, ..., e_C] \quad \text{s.t.} \quad e_j = \frac{T^o_j - t^o_j}{t_{\text{max}}}
\label{eq:error}
\end{equation}
where $t^o_j$ and $T^o_j$ are the actual and the target firing times of the $j$th output neuron, respectively.
Let us define $t_{\text{min}}$ as the minimum firing time in the output layer (i.e., $t_{\text{min}} = \min\{t^o_j \mid 1 \leq j \leq C\}$).
For an input image belonging to the $i$th category, we have
\begin{equation}
T^o_j =
\begin{cases}
t_{\text{min}} - \gamma & \text{if } j = i \\
t_{\text{min}} + \gamma & \text{if } j \neq i \land t^o_j < t_{\text{min}} + \gamma \\
t^o_j & \text{if } j \neq i \land t^o_j \geq t_{\text{min}} + \gamma
\end{cases}
\end{equation}

In a special case that all the output neurons remain silent during the forward pass (emit fake spikes at $t_{\text{max}}$), we set $T^o_i = t_{\text{max}} - \gamma$ and $T^o_{j \neq i} = t_{\text{max}}$ to force the correct neuron to fire.

Let us define the squared error loss function as
\begin{equation}
L = \frac{1}{2} \|e\|^2 = \frac{1}{2} \sum_{j=1}^C e_j^2
\end{equation}

To apply the gradient descent algorithm, we should compute $\frac{\partial L}{\partial W^l_{ji}}$, the gradient of the loss function with respect to the weights.
And, we can update the real-valued weights as
\begin{equation}
W^l_{ji} = W^l_{ji} - \eta \cdot \frac{\partial L}{\partial W^l_{ji}}
\end{equation}

In our implementation, we approximate the derivative $\frac{\partial t^l_j}{\partial W^l_{ji}}$ by a constant value whenever the presynaptic neuron spikes before the postsynaptic neuron, and zero otherwise. This reflects the temporal dependency of spike arrival on weight contributions.

\begin{equation}
\frac{\partial L}{\partial W^l_{ji}} =
\begin{cases}
- \delta^l_j & \text{if } t^{l-1}_i < t^l_j \\
0 & \text{otherwise}
\end{cases}
\label{eq:update}
\end{equation}

For the output layer ($l = o$) we have
\begin{equation}
\delta^o_j = \frac{\partial L}{\partial e_j} \cdot \frac{\partial e_j}{\partial t^o_j} = - \frac{e_j}{t_{\text{max}}}
\label{eq:delta}
\end{equation}

For the hidden layers ($l \neq o$), according to the backpropagation algorithm, we compute the weighted sum of the delta values of neurons in the following layer,
\begin{equation}
\delta^l_j = \sum_k \delta^{l+1}_k \cdot W^{l+1}_{kj} \cdot \mathbb{I}[t^l_j < t^{l+1}_k]
\label{Eq_delta}
\end{equation}

To simplify gradient computation, we use a surrogate derivative approach where the change in spike time with respect to membrane potential is approximated as a constant.
We approximate $\frac{\partial t^{l+1}_k}{\partial V^{l+1}_k} = -1$ if and only if $t^l_j \leq t^{l+1}_k$.

As shown in equation \ref{Eq_delta}, to propagate the error to the primary layers, we should calculate the
weighted sum of the delta values of neurons in the
subsequent layer, which has a high computation
load due to the multiplying two floating-point parameters (weights and deltas). Here, we convert
delta values of neurons to spike times to simplify
the computation of backpropagating error in equation  \ref{Eq_delta}. In the next section we explain the details
of the proposed method.

\subsubsection{Spike Time Backpropagation}

\begin{figure}[!t]
\centering
\includegraphics[width=\linewidth]{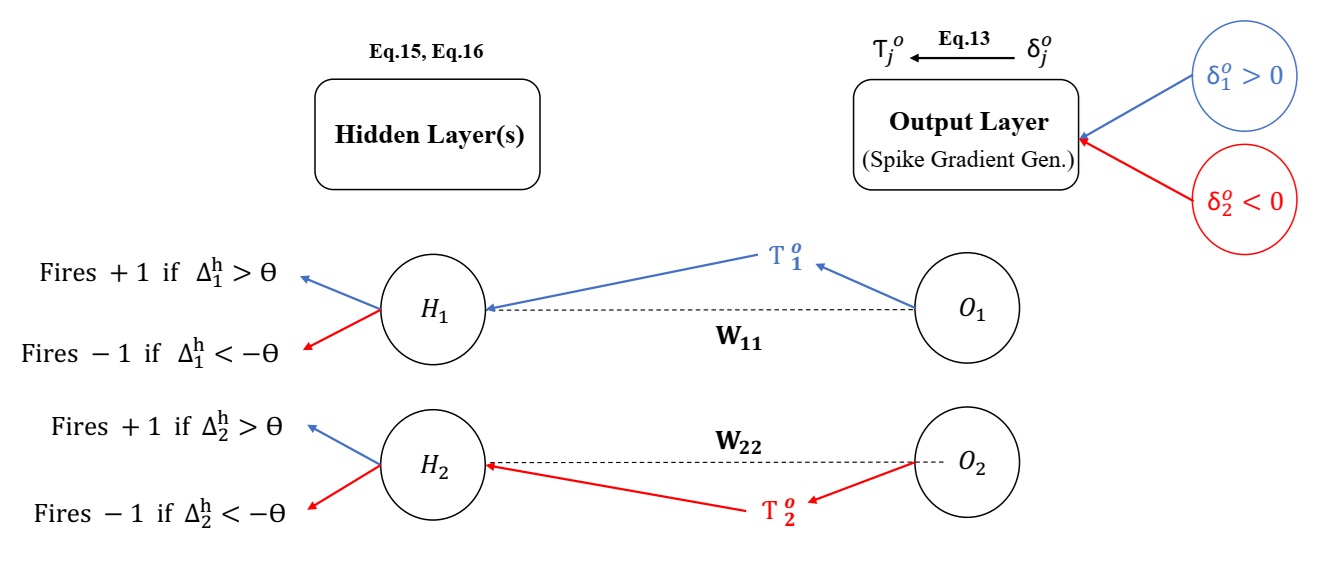}
\caption{Spike-based gradient propagation: \textbf{Blue} indicates positive gradients, \textbf{red} indicates negative gradients. Each hidden neuron fires a signed backward spike based on the sign of its delta.}
\label{backward_alg}
\end{figure}

In this section we explain the spike-based backpropagation. Figure~\ref{backward_alg} illustrates the proposed mechanism, where each neuron's gradient is encoded by a signed spike whose timing reflects its magnitude and whose polarity reflects its sign.
For the output layers, delta values are computed the same as equation~\ref{eq:delta}. But, for hidden layers, instead of computing equation~\ref{Eq_delta}, we want to backpropagate delta values using backward spike times. We convert the value of $\delta^o_j$ to the backward spike time $\tau^o_j$ (i.e. $\delta^o_j \rightarrow \tau^o_j$) for each output neuron $j$ (here $\tau$ corresponds to time in backward pass). In fact, the backward spike time of each neuron carries the information of delta value of that neuron. For positive delta, we have positive backward spike and, for negative value of delta, we send negative backward spike.

Before converting the delta values $\delta^o_j$ into backward spike times $\tau^o_j$, we first normalize them as $\delta^o_j = \delta^o_j / \sum_i \delta^o_i$, to prevent exploding or vanishing gradients. The normalized delta is then linearly mapped to the backward spike time as follows:
\begin{equation}
d^o_i = \text{round}(\delta^o_i \times t_{\text{max}})
\end{equation}

\begin{equation}
\tau^o_i =
\begin{cases}
t_{\text{max}} - d^o_i & \text{if } d^o_i > 0 \\
t_{\text{max}} + d^o_i & \text{if } d^o_i < 0
\end{cases}
\label{eq:FTP1}
\end{equation}

\begin{equation}
\mathbb{S}^o_i(\tau) =
\begin{cases}
1 & \text{if } \tau = \tau^o_i, \, d^o_i > 0 \\
-1 & \text{if } \tau = \tau^o_i, \, d^o_i < 0 \\
0 & \text{otherwise}
\end{cases}
\label{eq:FTP2}
\end{equation}

$\mathbb{S}^o_j(\tau)$ is the backward spike train from the output neuron $i$ which includes the information of the delta value of that neuron. Then, we backpropagate the error to hidden layers $l$ ($l \neq o$) by calculating the backward membrane potential of each neuron $i$ in layer $l$ ($\Delta^l_i$). $\Delta^l_i$ approximates $\delta^l_i$ value of the neuron $i$ by accumulating spike gradients.

\begin{equation}
\Delta(\tau)^l_i = \sum_j \mathbb{S}^{l+1}_j(\tau) W^{l+1}_{ji} [\tau^l_i < \tau^{l+1}_j]
\label{eq:back}
\end{equation}

Hence, we use the equation~\ref{eq:back} instead of equation~\ref{Eq_delta} to backpropagate the gradients. The IF neuron $i$ fires a backward spike only once the first time its backward membrane potential $\Delta^l_i$ crosses the threshold $\theta^l_i$ (i.e. $+\theta^l_i$ for positive backward spikes and $-\theta^l_i$ for negative backward spikes)

\begin{equation}
\mathbb{S}^l_i(\tau) =
\begin{cases}
1 & \text{if } \Delta^l_i(\tau) > \theta^l_i \\
-1 & \text{if } \Delta^l_i(\tau) < -\theta^l_i \\
0 & \text{otherwise}
\end{cases}
\label{eq:V_back}
\end{equation}

Note that we have temporal coding in the backward pass. In other words, earlier backward spikes correspond to larger gradients. The backward spike train $\mathbb{S}^l_i(t)$ indicates the delta value of neuron $i$ and it should be backpropagated through the network.

To update the hidden synaptic weights, we employ the normalized value of $\Delta^l_j$ as the delta value of neuron $j$ in equation~\ref{eq:update} ($\delta^l_j$). 

\section{RTL Implementation on FPGA}

The proposed neuromorphic hardware implements forward and backward passes of an SNN for the $8\times8$ Digits dataset. It processes 64 input pixels (0–15) using a three-layer network: two hidden layers of 20 neurons and an output layer of 10 neurons. Input spikes, latency-coded with 4-bit timestamps, are stored in memory and sent to the FPGA. Neurons that do not fire receive a virtual spike at step 15. Synaptic weights are 12-bit fixed-point (Q5.7) in BRAM, enabling multiplier-free, adder-based computations for low-power, efficient inference and learning.

\subsection{Hardware Implementation of the Forward Pass}
The forward pass, encompassing input encoding, spike propagation through hidden layers, and classification in the output layer, is realized through a modular, event-driven architecture. The design leverages the algorithm’s single-spike temporal coding, described in Section 3, to eliminate floating-point multiplications, ensuring a low-power and resource-efficient implementation suitable for neuromorphic edge applications.

\subsubsection{Key Hardware Modules}
The forward pass is implemented using three primary modules, each tailored to support the temporal dynamics of the proposed SNN algorithm:
\begin{itemize}
    \item \textbf{Integrate-and-Fire (IF) Neuron Module}: This module implements a non-leaky Integrate-and-Fire (IF) neuron, optimized for parallel processing as depicted in Figure~\ref{if_neuron}. Each neuron is connected to a dedicated Weight BRAM, storing 48-bit data blocks where each block contains four 12-bit fixed-point weights (Q5.7 format), enabling simultaneous retrieval of four weights per memory address. The neuron processes inputs over multiple cycles, handling four input weights and corresponding spike times concurrently per cycle. A 16-bit spike time input, comprising four 4-bit segments, is split into individual spike times using a Spike Splitter module. Similarly, the 48-bit weight data is divided into four 12-bit weights via a Weight Splitter module. The adder tree then accumulates the selected weights based on matching time steps, feeding the result to a Threshold Comparator to generate an output spike. The spike time output of each neuron is stored in a dedicated BRAM for later use in the backward phase.
    \begin{figure}[!t]
    \centering
    \includegraphics[width=\linewidth]{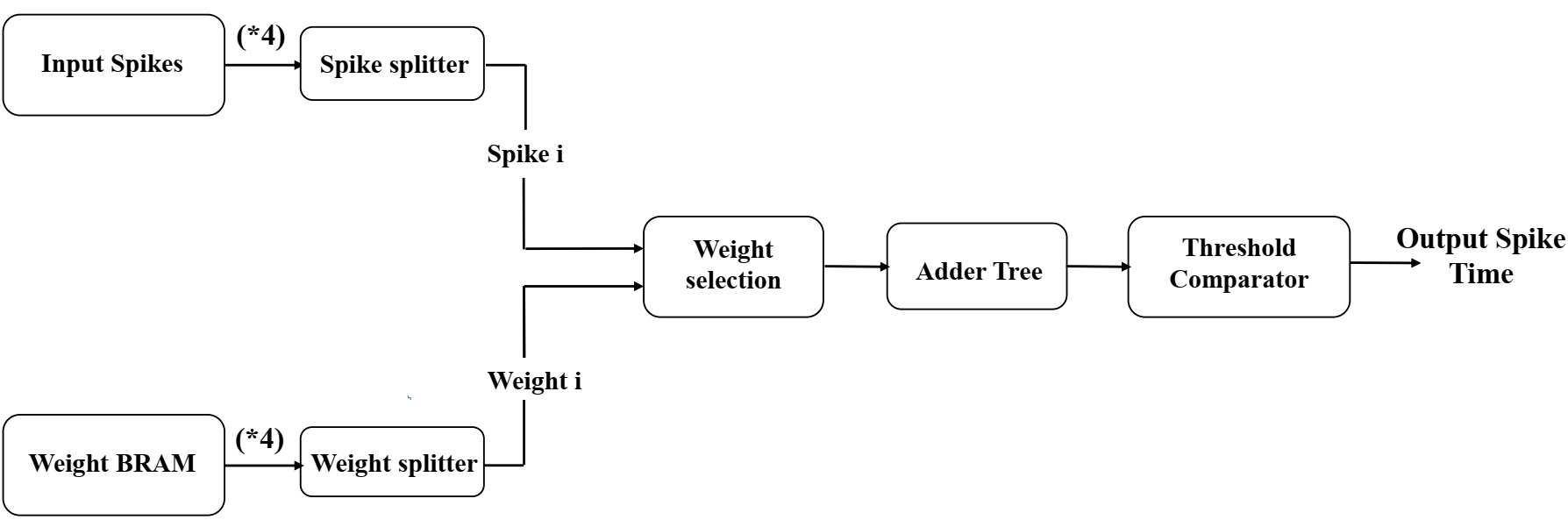}
    \caption{Schematic of a non-leaky IF neuron with parallel adder-based computations}
    \label{if_neuron}
    \end{figure}

    \item \textbf{Fully Connected Layer Module}: The fully connected layer module instantiates multiple IF neurons to process input spikes in parallel (Figure~\ref{FC_Layer}). The first hidden layer, comprising 20 neurons, connects to 64 inputs, completing processing over 16 clock cycles due to the parallel handling of four inputs per cycle. The second hidden layer, with 20 neurons and 20 inputs, completes processing in 5 clock cycles as each neuron processes four inputs concurrently per cycle. The module employs Universal Time Coding to manage spike propagation and weight storage, transmitting 4-bit spike timestamps between layers for efficient communication. Parameterized inputs enable flexibility in layer size and weight precision, enhancing scalability across FPGA platforms.
    \begin{figure}[!t]
    \centering
    \includegraphics[width=\linewidth]{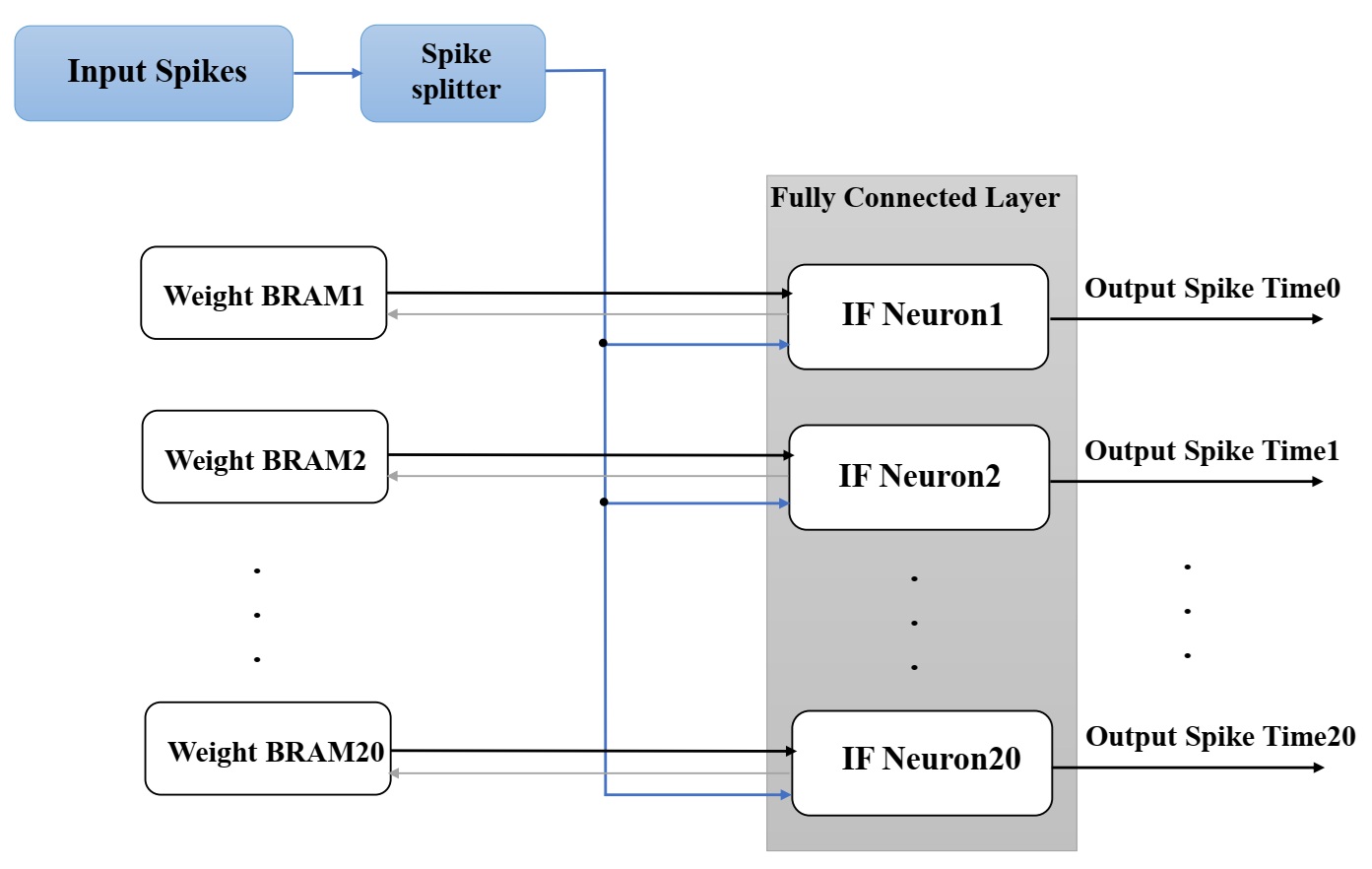}
    \caption{Fully connected layer with parallel IF neurons}
    \label{FC_Layer}
    \end{figure}

    \item \textbf{Output Layer Module}: The output layer module processes spikes from the final hidden layer over five clock cycles to produce classification decisions for the $8\times8$ Digits dataset. Comprising 10 neurons, each corresponding to a class label, the module identifies the neuron with the earliest spike time as the predicted class. A dedicated decision-maker logic compares spike timestamps, enabling low-latency classification without complex arithmetic operations.
\end{itemize}

\subsubsection{FPGA Implementation Details}

The forward pass leverages the FPGA’s parallel processing capabilities to compute spike propagation across all neurons in a layer, with processing completed over multiple clock cycles based on the number of inputs (e.g., 16 cycles for 64 inputs and 5 cycles for 20 inputs). The architecture exclusively uses LUTs for arithmetic operations and Flip-Flops for temporary storage, while weights and spike times are stored in dedicated Block RAM (BRAM) for each neuron, optimizing resource usage by avoiding DSP slices. Each Weight BRAM stores 48-bit blocks containing four 12-bit weights, enabling parallel retrieval per cycle, while spike time outputs are saved in separate BRAMs for the backward phase.

The design employs Universal Time Coding to manage spike propagation, encoding active spikes with 4-bit timestamps that are processed only when matching the current time step. This sparse, event-driven approach reduces interconnect bandwidth by transmitting only active spike events. A 4-bit digital counter increments through a 16-unit time window in single clock cycle steps, synchronizing spike processing across all neurons. The 16-unit temporal resolution balances efficiency with precise spike timing for the $8\times8$ Digits dataset, supporting accurate classification. Input spikes are retrieved via a memory-mapped interface, encoding the dataset into 64$\times$4-bit spike trains.

Parallel processing and Universal Time Coding-based communication support on-chip training by minimizing latency in the forward pass, enabling rapid weight updates through dedicated write enable signals. The absence of floating-point multipliers and the event-driven spike propagation significantly reduce latency and power consumption, making the architecture suitable for energy-constrained embedded systems.

\subsection{Hardware Implementation of the Backward Pass}
The backward pass implements spike-time-based backpropagation (Section 3) for on-chip supervised learning, converting gradients into spike timings. This adder- and bit-shift-based approach minimizes floating-point operations, enabling a near-multiplication-free, event-driven process for FPGA. It covers error computation at the output, gradient propagation to hidden layers, and weight updates across all layers.

\subsubsection{Key Hardware Modules}
As illustrated in Figure{\ref{backward_flows}}, the backward pass is implemented using four main modules, each designed to support spike-time-based gradient computation and weight updates, optimized for FPGA deployment with minimal multiplication operations:
\begin{figure}[!t]
\centering
\includegraphics[width=\linewidth]{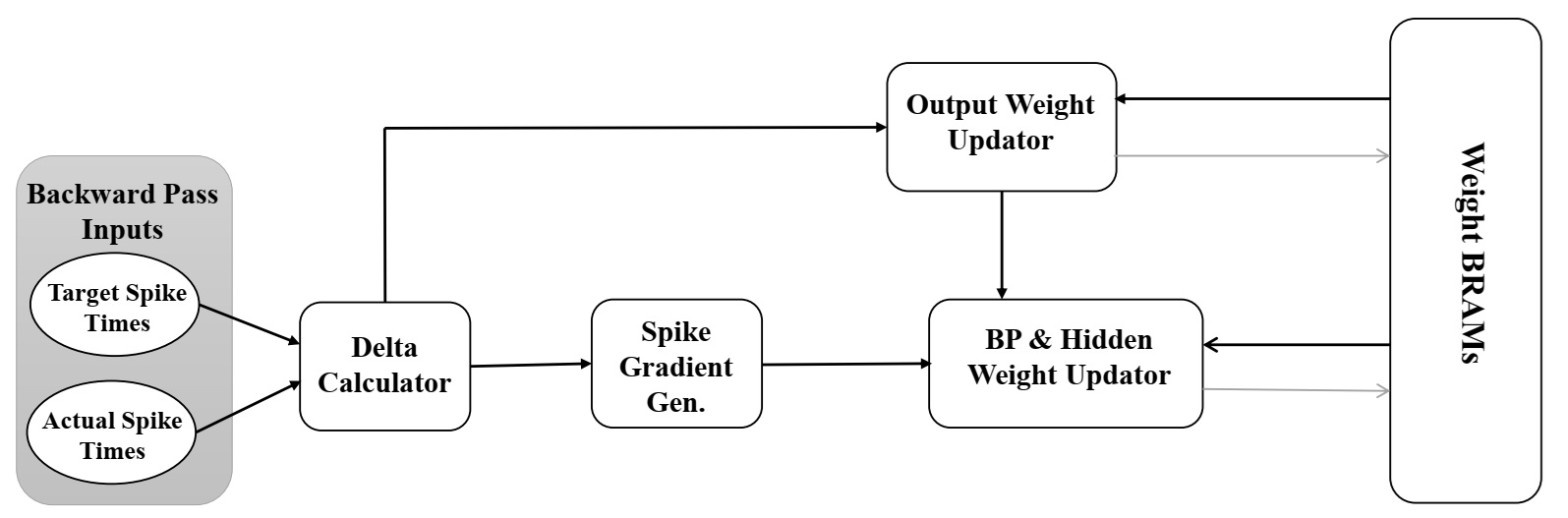}
\caption{Spike-based backward pass architecture with four key hardware modules.}
\label{backward_flows}
\end{figure}

\begin{itemize}
    \item \textbf{Delta Calculator Module}: This module computes the temporal error for the output layer by calculating the difference between target and actual spike times, as defined in Equation~(\ref{eq:error}). The errors are converted to 10-bit fixed-point format (Q1.9) using bit-shift and division operations, avoiding floating-point arithmetic.
    
    \item \textbf{Spike Gradient Generator Module}: This module transforms output layer deltas into positive (STP) and negative (STN) spike trains, as outlined in Equations~(\ref{eq:FTP1}) and~(\ref{eq:FTP2}), to align the backward phase with the forward phase’s event-driven structure. Utilizing simple bit-shift and comparison operations, it maps delta values to 5-bit spike times (including a sign bit to distinguish positive and negative spikes) within a 16-unit time window, facilitating sparse, event-driven gradient propagation to hidden layers without the need for complex multipliers.
    
    \item \textbf{Output Weight Updater Module}: This module updates the weights of the output layer based on the fixed-point deltas, input and output spike times (retrieved from BRAMs storing forward phase spike outputs), and a 10-bit fixed-point learning rate. It employs bit-shift operations for efficient gradient computation, with minimal use of multiplication limited to a single scalar product (delta × learning rate) that is subsequently downscaled via bit-shifting to fit the 12-bit weight format (Q4.8). The temporal dependency condition ($t_{\text{in}} < t_{\text{out}}$) ensures that weight updates occur only for synapses where the presynaptic spike precedes the postsynaptic spike, reflecting spike-timing-based gradient computation. This selective updating, combined with the near-multiplication-free approach, reduces computational overhead in the backward phase, enhancing hardware efficiency.

    \item \textbf{Backpropagation and Hidden Weight Updater Module}: This module computes gradients and updates hidden layer weights by propagating positive (STP) and negative (STN) spike gradients. Parallel neurons use a hardware architecture similar to the forward pass, sharing adder trees and BRAMs for weight retrieval and processing 4 spikes with 4 weights concurrently, only for synapses where the presynaptic spike precedes the postsynaptic spike. Each neuron retrieves a 48-bit BRAM block, splits it into four 12-bit weights, and feeds them to a shared adder tree; the accumulated potential is compared against a threshold to generate output spikes. Inactive neurons produce zero delta and an output spike time of 15. A sub-module computes deltas and updates weights in a single clock cycle using bit-shift operations with minimal multiplication, enabling a near-multiplication-free, event-driven learning mechanism optimized for FPGA.
    \end{itemize}

\subsubsection{FPGA Implementation Details}

\begin{figure}[!t]
\centering
\includegraphics[width=\linewidth]{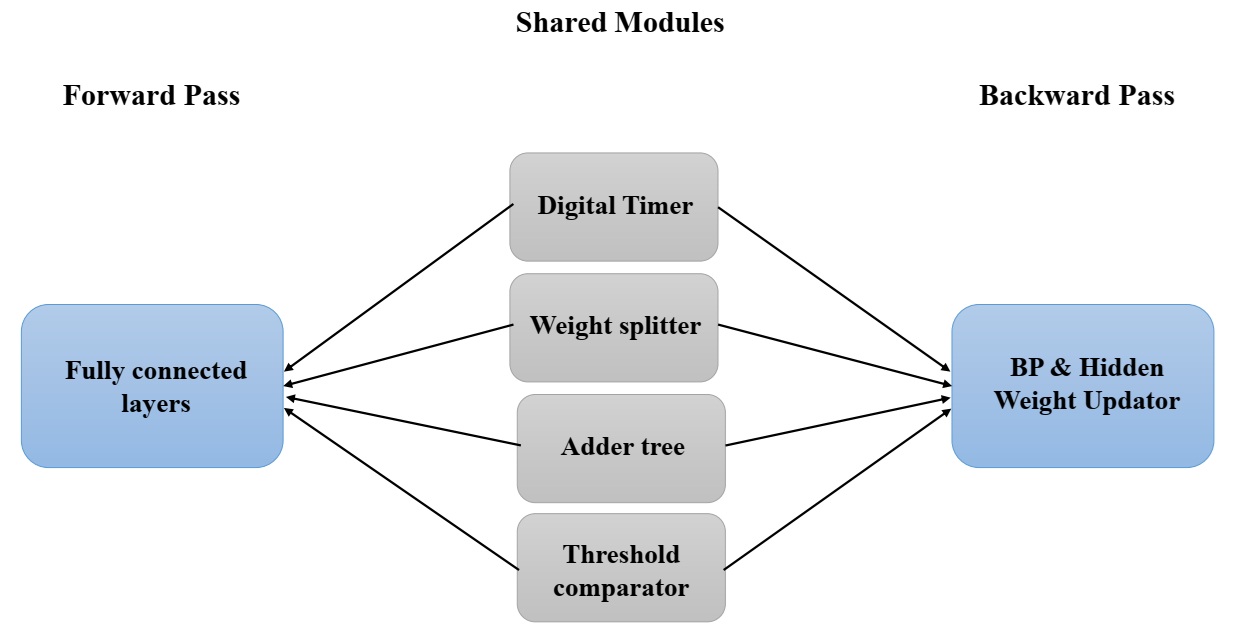}
\caption{Semi-shared FPGA architecture, optimizing resource usage and scalability.}
\label{shared_resources}
\end{figure}

The backward pass leverages FPGA parallelism to compute gradients and update weights across all neurons in 3–5 clock cycles per layer. As shown in Figure~\ref{shared_resources}, a semi-shared hardware design reuses adder trees, comparator/splitter units, BRAMs, and a 4-bit digital counter across forward and backward passes. Positive (STP) and negative (STN) spike gradients are handled by adjusting weights via two’s complement when spikes are negative, while core components remain shared to reduce resource usage. Gradients are converted into spike timings using adder-based computations and a single bit-shifted scalar multiplication, enabling a near-multiplication-free process.

Universal Time Coding transmits backward spikes as 5-bit timestamps only when active, minimizing interconnect bandwidth and power. Weights (12-bit) and deltas (10-bit) are fixed-point, and arithmetic operations are designed to be predominantly multiplication-free through bit-shift, comparison, and addition techniques, particularly leveraging addition for gradient propagation in the backward pass.

\section{Evaluation and Results}

We evaluate the proposed supervised spike-based training algorithm on MNIST, Fashion-MNIST, and 8×8 Digits datasets. The approach uses single-spike encoding and a multiplication-free learning rule, eliminating multipliers and reducing hardware complexity. Our goals are: (1) to demonstrate SNN classification performance under hardware-friendly constraints, (2) to validate performance on the 8×8 Digits dataset as a preliminary step for hardware implementation, and (3) to assess FPGA suitability, highlighting efficiency gains from the multiplier-free design. Evaluation includes algorithmic/software results (classification accuracy) and hardware results (resource utilization, maximum frequency, and throughput on Xilinx Artix-7). On the 8×8 Digits dataset, the method achieves 98

\subsection{Algorithmic Performance}
We evaluated our learning algorithm on MNIST, Fashion-MNIST, and 8×8 Digits datasets. For MNIST, two architectures were used: two-layer (784-400-10) and three-layer (784-400-400-10). For Fashion-MNIST (28×28 grayscale images), a three-layer network (784-600-600-10) achieved 84.8\% accuracy. The 8×8 Digits dataset, with 8×8 pixel images, was used to assess performance in a simpler, hardware-oriented setting. Single-spike encoding and the multiplication-free learning rule ensure compatibility with resource-constrained hardware while maintaining high accuracy. Table \ref{tab:accuracy} summarizes classification accuracies compared with leading SNN training algorithms, S4NN, STiDi-BP, BS4NN, Mostafa \cite{p9}, and HaSiST \cite{hasist}.

\begin{table*}[t]
\centering
\caption{Classification accuracy comparison of different SNN training methods on MNIST, Fashion-MNIST, and 8×8 Digits datasets.}
\label{tab:accuracy}
\begin{tabular}{|l||l||c||c||c||c||c|}
\hline
\textbf{Method} &  \textbf{Dataset} &  \textbf{Architecture} & \textbf{Accuracy (\%)} & \textbf{Coding scheme}& \textbf{Multiplication-Free} \\
\hline
S4NN \cite{p15} & MNIST & 784--400--10 & 97.4  & Single-Spike & \texttimes \\
S4NN \cite{p15} & Fashion-MNIST & 784--1000--10 & 88  & Single-Spike & \texttimes \\
STiDi-BP \cite{p17} & MNIST & 784--400--10 & 97.4  & Single-Spike & \texttimes \\
BS4NN \cite{p16} & MNIST & 784--600--10 & 97.0  & Single-Spike & \texttimes \\
BS4NN \cite{p16} & Fashion-MNIST & 784--1000--10 & 87.3  & Single-Spike & \texttimes \\
Mostafa \cite{p9} & MNIST & 784--800--10 & 97.2  & Single-Spike & \texttimes \\
HaSiST \cite{hasist} & MNIST & 784--200--10 & 97.5  & rate-based & \texttimes \\
HaSiST \cite{hasist} & 8*8 Digits & 64--20--10 & 99.0  & rate-based & \texttimes \\
Ours & MNIST & 784--400--10 & 97.4  & Single-Spike & $\checkmark$  \\
Ours & 8*8 Digits & 64--20--10 & 98.5  & Single-Spike & $\checkmark$ \\
S4NN \cite{p15} & MNIST & 784--400--400--10 & 96.7  & Single-Spike & \texttimes \\
STiDi-BP \cite{p17} & MNIST & 784--400--400--10 & 96.8  & Single-Spike & \texttimes \\
BS4NN \cite{p16} & MNIST & 784--400--400--10 & 96.5  & Single-Spike & \texttimes \\
BS4NN \cite{p16} & Fashion-MNIST & 784--600--600--10 & 86.2 & Single-Spike & \texttimes \\
Mostafa \cite{p9} & MNIST & 784--400--400--10 & 96.5  & Single-Spike & \texttimes \\
HaSiST \cite{hasist} & MNIST & 784--400--400--10 & 96.8  & rate-based & \texttimes \\
HaSiST \cite{hasist} & 8*8 Digits & 64--20--20--10 & 98.7  & rate-based & \texttimes \\
Ours & MNIST & 784--400--400--10 & 96.5  & Single-Spike & $\checkmark$ \\
Ours & Fashion-MNIST & 784--600--600--10 & 84.8  & Single-Spike & $\checkmark$ \\
Ours & 8*8 Digits & 64--20--20--10 & 98.0  & Single-Spike & $\checkmark$ \\
\hline
\end{tabular}
\end{table*}

In the two-layer architecture (784--400--10), our method achieves 97.4\% on MNIST, matching S4NN~\cite{p15}, while HaSiST~\cite{hasist} reaches 97.5\% on a smaller two-layer network (784--200--10). Unlike S4NN, STiDi-BP, BS4NN, Mostafa, and HaSiST, which rely on floating-point multiplications and gradient storage, our approach propagates gradients through spike timing, fully compatible with hardware constraints.
In the deeper three-layer architecture (784--400--400--10), our method achieves 96.5\% accuracy on MNIST, which is highly competitive with benchmark methods: S4NN (96.7\%), STiDi-BP (96.8\%), BS4NN (96.5\%), Mostafa (96.5\%), and HaSiST (96.8\%).
For Fashion-MNIST, using a three-layer architecture (784--600--600--10), our method achieves 84.8\% accuracy, closely approaching S4NN (88\%) and BS4NN (87.3\%) with larger two-layer architectures (784--1000--10). On the 8×8 Digits dataset (64--20--20--10), we achieve 98.0\%, competitive with HaSiST (98.7\%).

These results demonstrate the scalability and generalization of our multiplication-free, single-spike, fixed-point method, delivering near state-of-the-art performance while eliminating floating-point operations, making it highly suitable for low-power, resource-constrained neuromorphic hardware.

\subsection{Sparsity Analysis}
To further highlight the computational efficiency of our approach, we analyzed the mean number of active synapses in the forward pass for the 784--400--400--10 architecture on the MNIST dataset. Table \ref{tab:active_synapses} reports the number of active synapses per class. On average, only approximately 75\% of the total 477600 synapses are active during the forward pass, compared to traditional ANNs where all synapses contribute to computations. This sparsity significantly reduces the number of operations, making our method highly suitable for neuromorphic hardware, where energy efficiency and resource utilization are critical.

\begin{table*}[ht]
\centering
\caption{The number of Active Synapses and Their Percentage per Class for a sparse 784--400--400--10 SNN on MNIST}
\label{tab:active_synapses}
\resizebox{\textwidth}{!}{%
\begin{tabular}{|l||c||c||c||c||c||c||c||c||c||c|}
\hline
 \textbf{Class number} & \textbf{0} & \textbf{1} & \textbf{2} & \textbf{3} & \textbf{4} & \textbf{5} & \textbf{6} & \textbf{7} & \textbf{8} & \textbf{9} \\
\hline
\textbf{Active Synapses} & 403750 & 268196 & 377422 & 374153 & 348051 & 376316 & 366351 & 337356 & 383874 & 346152 \\
\hline
\textbf{Percentage (\%)} & 84.54 & 56.16 & 79.03 & 78.34 & 72.87 & 78.80 & 76.71 & 70.64 & 80.37 & 72.48 \\
\hline
\end{tabular}%
}
\end{table*}

\subsection{Hardware Implementation Results}

Our method overcomes challenges of traditional gradient-based SNN training, which rely on floating-point arithmetic, gradient storage, and centralized computation, by using single-spike encoding and a multiplication-free learning rule. Gradients are encoded into backward spike latencies and processed with lightweight, integer-based, event-driven operations. Each neuron integrates signed backward spikes ($\pm1$) and fires when its backward potential crosses a threshold, mirroring forward IF dynamics without floating-point operations or gradient storage. Weights are 12-bit fixed-point, balancing precision and efficiency.

The single-spike constraint ensures ultra-sparse communication, predictable timing, and reduced energy consumption, critical for edge neuromorphic applications. Unlike HaSiST and SG-based methods, our architecture supports distributed, asynchronous computation with independent weight updates per layer, eliminating centralized coordination and BPTT-like unfolding. Table~\ref{tab:comp_complexity} summarizes key computational characteristics, highlighting the suitability of our approach for resource-constrained FPGA implementation. Quantitative results for resource utilization, power, and throughput are presented in the following subsections.

\begin{table*}[h!]
    \centering
    \caption{Qualitative comparison of computational characteristics of SNN training methods for on-chip learning.}
    \label{tab:comp_complexity}
    \resizebox{\textwidth}{!}{%
    \begin{tabular}{|l||c||c||c||c||c||c|}
        \hline
        \textbf{Feature / Method} & \textbf{Mostafa~\cite{p9}} & \textbf{STiDi-BP~\cite{p17}} & \textbf{BS4NN~\cite{p16}} & \textbf{HaSiST~\cite{hasist}} & \textbf{SG-Based~\cite{SG1}} & \textbf{Ours} \\
        \hline
        Floating-point multiplications & High & High & Moderate & Moderate & Very High & None \\
        Additions (per layer)         & Moderate & Moderate & Moderate & Moderate & High & Moderate \\
        Gradient storage             & Required & Required & Partial & Partial & Required & Not required \\
        Spike-time comparisons       & Required & Required & Required & Required & Not used & Used only \\
        Weight type                 & Full-precision & Full-precision & Binary & Integer (8-bit) & Full-precision & Fixed-point \\
        Backprop signal             & Float deltas & Float deltas & Float deltas & Sigmoid-based & Surrogate deltas & Spike latencies \\
        Gradient computation unit   & Shared, float-based & Shared, float-based & Shared, float-based & Shared, sigmoid-based & Global, float-based (BPTT) & Distributed, integer-based \\
        Neuron operations           & Accumulate \& fire & Accumulate \& fire & Accumulate \& fire & LIF (forward), Sigmoid (backward) & Leaky IF, multi-spike & IF + backward IF \\
        Spike per neuron            & 1 & 1 & 1 & Many & Many & 1 \\
        Memory bandwidth            & High & High & Moderate & Moderate & Very High & Low \\
        \hline
    \end{tabular}%
    }
\end{table*}

\subsubsection{Resource Utilization}
To evaluate the hardware efficiency of the proposed spiking neural network (SNN) design, we report the FPGA resource utilization for both the forward and backward passes, including slice registers and look-up tables (LUTs). The evaluation targets a compact 64--20--20--10 architecture (1,880 synapses) for the 8×8 Digits dataset, which is well-suited for resource-constrained applications due to its low synapse count and compatibility with edge devices.

Our design leverages single-spike encoding and a multiplication-free learning rule, completely eliminating the need for DSP slices and minimizing switching activity through event-driven integer operations. This approach significantly reduces hardware complexity compared to prior works, such as HaSiST~\cite{hasist}, which employs rate-based encoding and floating-point operations for both forward and backward passes. Table~\ref{tab:resource_utilization} compares the resource utilization of our forward and backward passes with HaSiST, highlighting the efficiency gains achieved through our multiplier-free design.

For the forward pass, our design utilizes 1,256 slice registers and 3,190 LUTs, corresponding to approximately 0.66 slice registers and 1.68 LUTs per synapse. These figures reflect a substantial reduction compared to HaSiST, which requires 1.03 slice registers and 2.80 LUTs per synapse for its forward pass. The complete absence of DSP slices and the low resource consumption in our forward pass stem from relying solely on spike-time computation and fixed-point addition, enhancing hardware efficiency. By employing 12-bit fixed-point weights, our approach achieves a balance between computational precision and resource utilization, outperforming HaSiST’s 8-bit integer weights, which may limit accuracy in certain scenarios, and other methods like Mostafa~\cite{p9} and STiDi-BP, which depend on full-precision synaptic weights.

For the backward pass, our design maintains the multiplication-free philosophy, encoding gradient information into backward spike latencies and processing them using lightweight, event-driven fixed point operations. This eliminates the need for explicit gradient storage and complex floating-point computations, unlike HaSiST, which requires partial gradient storage and sigmoid-based operations for the backward pass. Our backward pass utilizes approximately 1.33 slice registers and 5.1 LUTs per synapse, significantly outperforming HaSiST’s backward pass (2.63 slice registers and 37.84 LUTs per synapse)~\cite{hasist}, with an approximately 7.4-fold reduction in LUT usage and a 2-fold reduction in register usage. The absence of DSP slices underscores the efficiency of our distributed, asynchronous weight update mechanism.

The ultra-low resource usage of both forward and backward passes, combined with distributed and asynchronous computation, which is enabled by single-spike encoding and multiplication-free spike-time-based backpropagation, significantly enhances our design’s efficiency and potential scalability for larger networks. By eliminating the need for centralized coordination and complex gradient computations (e.g., BPTT or surrogate gradients), our approach minimizes resource overhead, making it an ideal candidate for on-chip forward and backward passes in low-power, resource-constrained neuromorphic hardware, such as edge devices.

\begin{table*}[ht]
\centering
\caption{FPGA resource utilization comparison for SNN implementations.}
\label{tab:resource_utilization}
\begin{tabular}{|l|c||c||c||c||c|}
\hline
\textbf{Method} & \textbf{Platform} & \multicolumn{2}{c|}{\textbf{Slice Registers/Synapse}} & \multicolumn{2}{c|}{\textbf{LUTs/Synapse}} \\
\cline{3-4} \cline{5-6}
& & \textbf{Forward} & \textbf{Backward} & \textbf{Forward} & \textbf{Backward} \\
\hline
HaSiST~\cite{hasist} & Virtex-6 & 1.03 & 2.63 & 2.8 & 37.84 \\
Ours & Artix-7 & 0.66 & 1.33 & 1.68 & 5.1 \\
\hline
\end{tabular}
\end{table*}
\begin{table*}[ht]
\centering
\caption{Performance comparison for forward and backward passes of SNN implementations.}
\label{tab:performance_comparison}
\begin{tabular}{|l||c||c||c||c|}
\hline
\textbf{Method} & \textbf{Platform} & \multicolumn{2}{c|}{\textbf{Fmax (MHz)}} & \textbf{Throughput (FeaPS)} \\
\cline{3-4}
& & \textbf{Forward} & \textbf{Backward} & \\
\hline
HaSiST~\cite{hasist} & Virtex-6 & 135.073 & 50 & $0.03 \times 10^9$ \\
Ours & Artix-7 & 125.5 & 105.3 & $0.0316 \times 10^9$ \\
\hline
\end{tabular}
\end{table*}

\subsubsection{Maximum Frequency and Throughput}
 This section presents the \textbf{maximum frequency} and \textbf{throughput} for both forward and backward passes, with comparisons to HaSiST~\cite{hasist} on the Virtex 6 platform. A key feature of our design is the absence of floating-point multipliers in the backward pass, achieved by encoding gradients into spike latencies. This approach eliminates the need for complex multiplication operations, unlike HaSiST, which relies on distinct LIF neurons for inference and sigmoid-based neurons for training, thereby increasing resource demands.

\paragraph{Maximum Frequency (Fmax)}
Our design achieves a maximum frequency of \textbf{125.5 MHz} for the forward pass and an impressive \textbf{105.3 MHz} for the backward pass. While the forward pass operates at a solid frequency suitable for high-speed processing, the backward pass significantly surpasses HaSiST’s \textbf{50 MHz} training~\cite{hasist} frequency, representing a remarkable \textbf{110.6\%} improvement. This enhancement in the backward pass is driven by the absence of floating-point multipliers, relying solely on fixed-point additions and spike-time comparisons. The consistent design across passes, facilitated by shared modules, underscores the efficiency of our approach for high-speed neuromorphic processing on edge devices.

\paragraph{Throughput}
Throughput evaluates the real-time processing capability of our SNN and is measured in \textbf{features per second (FeaPS)}.
For a fair comparison with HaSiST’s two-layer design (64-20-10), we consider a two-layer architecture that also requires 21 cycles (16 for the first layer and 5 for the output layer).
Our design achieves $0.0316 \times 10^9$ FeaPS at 142.45~MHz, processing approximately 0.452 million samples per second.
Compared to HaSiST’s $0.03 \times 10^9$ FeaPS at 135.073~MHz~\cite{hasist}, our design offers a \textbf{5.47\%} higher throughput, enabled by efficient parallel processing within layers and a higher operating frequency.

\section{Conclusion}
This work presented a multiplication-free, spike-time-based supervised learning algorithm and its FPGA implementation, enabling efficient on-chip training of deep SNNs. The proposed framework achieves competitive accuracy on MNIST, Fashion-MNIST, and 8×8 Digits datasets while eliminating floating-point operations and maintaining sparse, event-driven computation.
Implemented on a Xilinx Artix-7 FPGA, the architecture achieves higher operating frequencies, improved throughput, and significantly lower resource utilization compared to prior designs such as HaSiST. By reusing computation and memory units across forward and backward passes, the system minimizes hardware redundancy and routing congestion, achieving an effective balance between speed, scalability, and resource cost.

While single-spike temporal encoding enhances sparsity and efficiency, inactive neurons still transmit default time packets, preventing fully event-driven operation. Future work will focus on spike-time quantization, adaptive synchronization, and architectural optimization to further improve energy efficiency and scalability for larger networks on resource-constrained FPGAs.


\bibliographystyle{IEEEtran}
\bibliography{references}

\end{document}